\documentclass{article}
\usepackage[utf8]{inputenc}
\usepackage[margin=1.4in]{geometry}
\usepackage{comment}
\usepackage{natbib}
\usepackage{booktabs}
\usepackage{url}

\usepackage{graphicx}
\usepackage{wrapfig}
\usepackage{multirow}
\usepackage{placeins}
\usepackage{amsmath}
\usepackage{amssymb}
\usepackage{bbm}

\allowdisplaybreaks

\newcommand{\prob}{\mathcal{P}}
\newcommand{\ent}{\mathcal{H}}
\newcommand{\uniq}{\mathcal{U}}
\newcommand{\reals}{\mathcal{R}}
\newcommand{\nats}{\mathcal{N}}
\newcommand{\ex}{\mathbb{E}}
\newcommand{\spa}{\hspace{0.25em}}

\newcommand{\df}{d}

\usepackage{amsthm}
\usepackage{cleveref} % order

\newtheorem{theorem}{Theorem}[section]
\newtheorem{definition}[theorem]{Definition}
\newtheorem{lemma}[theorem]{Lemma}
\newtheorem{corollary}[theorem]{Corollary}

\title{Properties of Minimizing Entropy}
\author{Xu Ji, L\'ena N\'ehale-Ezzine, Maksym Korablyov \\
Mila, Quebec AI Institute}

\begin{document}

\maketitle

\begin{abstract}
    Compact data representations are one approach for improving generalization of learned functions. We explicitly illustrate the relationship between entropy and cardinality, both measures of compactness, including how gradient descent on the former reduces the latter. Whereas entropy is distribution sensitive, cardinality is not. We propose a third compactness measure that is a compromise between the two: expected cardinality, or the expected number of unique states in any finite number of draws, which is more meaningful than standard cardinality as it discounts states with negligible probability mass. We show that minimizing entropy also minimizes expected cardinality.
\end{abstract}

\section{Introduction}
The compactness of data representations is a recurring theme in research on generalization in machine learning. 
Broadly speaking, given the problem of learning a mapping from input space $\mathcal{X}$ to output space $\mathcal{Y}$ from empirical samples of inputs and targets, and a choice over pairs of representation and inference functions $(g_i, f_i)$ where $g_i : \mathcal{X} \rightarrow \mathcal{Z}_i$ and $f_i: \mathcal{Z}_i \rightarrow \mathcal{Y}$, one prefers the instance of $(g_i, f_i)$ whose $\mathcal{Z}_i$ is most compact, all else equal, by some definition of compactness. %Synonyms of compact representations are simple or low-population representations.
%This intuition is related to Occam's Razor, or preferring the simplest inference function all else equal, because compactness of input space constrains the complexity of functions defined on it. 
%In the statistical study of generalization, guarantees on the deviation of test error from empirical training error for a learned function improve with compact input spaces and simpler function classes. A classical measure of the latter is the number of function parameters.

This intuition is suggested by generalization error bounds that scale with covering number or entropy \citep{xu2012robustness,bassily2018learners}, the information bottleneck principle  that advocates minimizing entropy of data representations \citep{tishby2000information}, discretization of continuous representations which reduces cardinality of the 0-width covering from infinite to finite \citep{liu2021discrete}, maximizing sparsity or minimizing overlap across representations more generally \citep{goyal2020inductive}, and estimating reliability of predictions using local density of training samples in representation space \citep{ji2021test,tack2020csi}.
It is also related to Occam's Razor generalization error bounds (review in \citet{mohri2018foundations}), or preferring the class of simplest inference functions all else equal, because compactness of a space constrains the complexity of functions defined on it. 

There are different measures of compactness of a state space. 
In particular there is a link between entropy, which is a distribution sensitive measure of compactness, and cardinality, which is distribution insensitive. Intuitively one would expect reducing entropy to monotonically reduce the number of states with non-zero probability, shrinking cardinality, and we show this explicitly for a generic model of unconstrained probability masses.
We consider a third measure of size that is not often used: the expected number of unique states in any finite $m$ draws of a state space. This measure corresponds to expected cardinality of finite sample draws and is tighter than the standard cardinality measure, being upper bounded by it. 
In highly skewed distributions, it is significantly lower, as states with non-zero probability are counted by cardinality even if the probability of drawing them is negligible. 
We show that expected cardinality is also reduced by minimizing entropy. 
We conclude that for optimization purposes, entropy can provide a convenient continuous surrogate for minimizing the cardinality of states in a learned data representation.

\vspace{-1em}
\begin{figure}[h]
    \centering
    \includegraphics[width=0.5\textwidth]{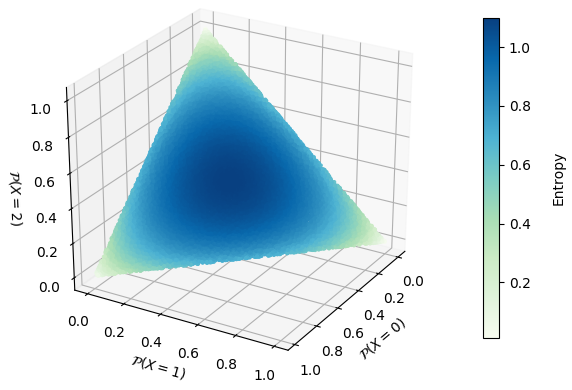}
    \caption{Entropy for 50K randomly sampled probability distributions of a discrete variable with 3 states. Entropy is maximized for uniform distributions (centre), minimized for one-hot distributions (corners). Lower entropy of any distribution $a$ compared to any distribution $b$ is not a sufficient condition for fewer non-zero states in $a$ (there exist darker points on the edges than on the interior), but as the color gradient implies, minimizing entropy via local optimization of probability mass is sufficient for decreasing the number of non-zero states across optimization timesteps.} % todo add legend
    \label{fig:entropies}
\end{figure}

\section{Discrete variable case}\label{s:properties}

Consider a vector of non-negative reals, $z \in \mathcal{R}_{\geq 0}^{s}$, which represents the unnormalized probability mass of a discrete variable $X$ taking each of $s$ states. Then the normalized probabilities are $\prob(X = i | z) = \frac{z_i}{\sum_{j} z_{j}}$ where $j \in [0, s)$ is left implicit for readability, and entropy is:
\begin{align}
    \ent(z) \triangleq - \sum_{i} \frac{z_i}{\sum_{j} z_{j}} \log  \frac{z_i}{\sum_{j} z_{j}}.
\end{align}
\begin{corollary}\label{thm:0}
\textbf{Gradient of entropy with respect to each probability mass.}
\begin{align}
\frac{\df \ent(z)}{\df z_{k}} &=   
\frac{\df}{\df z_{k}} \Bigg ( - \frac{z_k}{\sum_{j} z_{j}} \log  \frac{z_k}{\sum_{j} z_{j}}
- 
\sum_{i \neq k} \frac{z_i}{\sum_{j} z_{j}} \log  \frac{z_i}{\sum_{j} z_{j}} \Bigg ) \\
&= \frac{- \sum_{i \neq k } z_i }{ \big( \sum_{j} z_j \big)^2 } 
+ \frac{- \sum_{i \neq k } z_i }{ \big(\sum_{j} z_j \big)^2 } \log \frac{z_k}{ \sum_{j} z_j }
- 
\sum_{i \neq k} \bigg (  \frac{- z_i}{ \big( \sum_{j} z_j \big)^2} 
+ \frac{- z_i}{ \big( \sum_{j} z_j \big)^2} \log \frac{z_i}{ \sum_{j} z_j }\bigg ) \\
&= \frac{- \sum_{i \neq k } z_i }{ \big( \sum_{j} z_j \big)^2 } \Bigg ( 1 + \log \frac{z_k}{ \sum_{j} z_j } \Bigg ) 
- 
\sum_{i \neq k}  \frac{- z_i}{ \big( \sum_{j} z_j \big)^2} \bigg ( 1 + \log \frac{z_i}{ \sum_{j} z_j } \bigg )  \\
&= \frac{- \sum_{i \neq k } z_i }{ \big( \sum_{j} z_j \big)^2 } \Bigg (\log \frac{z_k}{ \sum_{j} z_j } \Bigg ) 
+ 
\sum_{i \neq k}  \frac{z_i}{ \big( \sum_{j} z_j \big)^2} \bigg ( \log \frac{z_i}{ \sum_{j} z_j } \bigg )  \\
&= 
\frac{1}{\big( \sum_{j} z_j \big)^2} \sum_{i \neq k} z_i \log \frac{z_i}{z_k}\\
&=  \frac{1}{\big( \sum_{j} z_j \big)^2} \sum_{i} z_i \log \frac{z_i}{z_k} \label{eq:h_grad} \\
&\bigg [ =\frac{1}{\big( \sum_{j} z_j \big)^2}  \sum_{i} z_i \bigg( \log \frac{z_i}{\sum_{j} z_j} - \log \frac{z_k}{\sum_{j} z_j} \bigg ) \label{eq:h_grad2} \\
&= \frac{1}{\sum_{j} z_j} \sum_i \prob(X = i | z) \log \frac{\prob(X = i | z)}{\prob(X = k)}
\bigg ] .
\end{align}
\end{corollary}
Consider taking one gradient step of minimizing entropy $\ent$ with respect to masses $z$, $\tilde{z}_k = \max(z_k - \eta \frac{d \ent}{d z_k}, 0)$, where $\eta$ is a learning rate hyperparameter.
The following results follow directly from the equation for $\frac{d \ent}{d z}$ in \cref{eq:h_grad}.

\begin{corollary} \label{thm:0b}
\textbf{State with the smallest non-zero mass always decreases in mass, and state with the largest mass always increases in mass.}
Follows directly from \cref{thm:0}; the smallest non-zero mass (and any zero masses) has a positive gradient, and the largest mass has a negative gradient.
\end{corollary}

\begin{corollary} \label{thm:1}
\textbf{Condition for whether a state's mass decreases or increases.}
A necessary and sufficient condition for the gradient to be positive (i.e. the mass should be decreased to decrease entropy) is the log mass of the state being less than the weighted average of all log masses:
\begin{align}
    \frac{\df \ent}{\df z_{k}} > 0 \longleftrightarrow 
\log z_k
< 
\sum_{i} \frac{z_i}{\sum_j z_j} \log z_i, \label{eq:dec_cond}
\end{align}
and vice versa for negative gradient or increase in mass.
\end{corollary}

\begin{corollary} \label{thm:2}
\textbf{Mass order is preserved and gaps are widened.} 
For all $k, k': z_k < z_{k'}$, the decrease in mass $z_k$ is larger than the decrease in mass $z_{k'}$, meaning gradients decrease monotonically going from the smallest to largest non-zero masses:
\begin{align}
    \eta\frac{\df \ent}{\df z_{k}} > \eta\frac{\df \ent}{\df z_{k'}} 
    \longleftrightarrow 
    \sum_{i} z_i \log z_k < \sum_{i} z_i \log z_{k'}
    \longleftrightarrow \top.
\end{align}
\end{corollary}

\begin{corollary} \label{thm:3}
\textbf{Unless the masses start as uniform, they are never uniform during entropy minimization.}
Follows directly from \cref{thm:2}.
\end{corollary}

\begin{corollary} \label{thm:4}
\textbf{Zero masses stick to zero.}
\Cref{eq:dec_cond} is always true if $z_k = 0$, but $z$ is defined on the non-negative reals, so states with probability 0 stay at probability 0. 
\end{corollary}

\begin{corollary}\label{thm:8}
\textbf{Sum of gradients of masses is positive.}
From \cref{app:1}, for any non-uniform $a \in \reals_{\geq 0}^s$, $\sum_j \sum_i a_i \log \frac{a_i}{a_j} > 0$. Therefore 
$\sum_k \frac{\df \ent(z)}{\df z_{k}} = \frac{1}{\big( \sum_{j} z_j \big)^2} \sum_k  \sum_{i} z_i \log \frac{z_i}{z_k} $ $> 0$.
This implies average mass gradient is positive, $\frac{1}{s} \sum_{k} \frac{\df \ent(z)}{\df z_{k}} > 0$, and total mass is guaranteed to decrease in any iteration where no gradients are clipped to maintain non-negative masses, $\sum_j \tilde{z}_j < \sum_j z_j $.
\end{corollary}

\begin{corollary}
 \label{thm:6}
\textbf{Probability order is preserved.}
This follows from \cref{thm:2}, since mass order is preserved and the new probabilities are the new masses divided by a value that is constant across all states.
\end{corollary}

\begin{corollary} \label{thm:5}
\textbf{Change in probability for each state}. Consider the case where the smallest non-zero mass state in $z$ still has positive mass in $\tilde{z}$ (otherwise, we know the number of non-zero probability states decreases). Since this state maintains positive mass, all states with greater mass in $z$ also maintain positive mass, by \cref{thm:2}, so none of their gradients are clipped in the optimization step.
Then for any non-zero mass state in $z$ with index $i$, its increase in probability is given by:
\begin{align}
&\frac{z_i - \frac{\eta}{(\sum_l z_l)^2} \sum_k z_k \log \frac{z_k}{z_i}}{\sum_j z_j - \frac{\eta}{(\sum_l z_l)^2} \sum_k z_k \log \frac{z_k}{z_j}} - 
\frac{z_i}{\sum_j z_j} \\
= &\frac{\eta}{c} \bigg [ 
\overbrace{z_i \sum_j \frac{1}{(\sum_l z_l)^2} \sum_k z_k \log \frac{z_k}{z_j}}^{\text{mass for $i$ $\times$ mass gradients for all $j$}}
- \overbrace{(\sum_j z_j) \frac{1}{(\sum_l z_l)^2} \sum_k  z_k \log \frac{z_k}{z_i}}^{\text{mass for all $j$ $\times$ mass gradient for $i$ }}
\bigg ], 
\end{align}
where $c = (\sum_j z_j) (\sum_j z_j - \frac{\eta}{(\sum_l z_l)^2} \sum_k z_k \log \frac{z_k}{z_j}) > 0$ is the same for all $i$. 
\end{corollary}

\begin{corollary} \label{thm:7}
\textbf{State with the smallest non-zero probability always decreases in probability, and state with the largest probability always increases in probability.}
From \cref{thm:5}, the increase in probability for state $i$ is $\frac{\eta}{c} > 0$ multiplied by:
\begin{align}
& \bigg (  z_i \sum_j \frac{1}{(\sum_l z_l)^2} \sum_k z_k \log \frac{z_k}{z_j} \bigg ) 
- \bigg (   (\sum_j z_j) \frac{1}{(\sum_l z_l)^2} \sum_k  z_k \log \frac{z_k}{z_i} \bigg )\\ 
&= s \spa z_i \spa \gamma
- s \spa \mu \spa \frac{\df \ent(z)}{\df z_{i}},
\end{align}
where $\gamma = \frac{1}{s} \sum_{k} \frac{\df \ent(z)}{\df z_{k}} > 0$ (\cref{thm:8}) and $\mu = \frac{1}{s} \sum_k z_k > 0$ and denote average gradient and mass. For the smallest non-zero mass, $0 < z_i < \mu$ and $0 < \frac{\df \ent(z)}{\df z_{i}} > \gamma$, and for the largest mass, $z_i > \mu > 0$ and $0 > \frac{\df \ent(z)}{\df z_{i}} < \gamma$.

%change in probability is positive for the largest old mass (which has the most negative mass gradient, \cref{thm:2}) and negative for the smallest old mass (which has the most positive mass gradient, \cref{thm:2}).
\end{corollary}

\begin{corollary}
\textbf{Uniform distribution uniquely maximises entropy, non-unique one-hot distributions minimize it.}
These are the cases where $\forall k: \tilde{z}_k - z_{k} = 0$: either when the masses are all equal so $\frac{d \ent}{d z} = \bar{0}$, or there is only one non-zero mass, in which case the non-zero mass has gradient 0 and the zero masses have gradient $\infty$, staying clamped at 0. 
%These are also when all the partial derivatives w.r.t. probabilities are 0. 
These critical points can also be found using Lagrangian constrained optimization (a tutorial is given in \citet{infotheory}).
\end{corollary}

\begin{theorem}\label{thm:entropy_cardinality}
\textbf{The decrease in the probability of the smallest non-zero mass is lower bounded by a value that is $> 0$ and increases with time, so applying gradient descent decreases the number of non-zero probability states in a finite number of iterations.}
More concretely, let $z^t$ be the vector of non-uniform masses at timestep $t$ and let the number of non-zero states in it be $c^t \triangleq \sum_{i=0}^{s-1} \mathbbm{1}_{z^{t}_i > 0}$. For each timestep $t \in \nats$ where $c^t > 1$, there exists finite $\tilde{t}$ where $c^{\tilde{t}} < c^t$, and therefore there exists finite $t^*$ such that $c^{t^*} = 1$.
\end{theorem}

We prove that at any time step $t$, the number of non-zero masses decreases given a finite number of subsequent optimization steps. In timestep $t$, assume the case where the state with smallest non-zero mass does not reach 0 mass in $t+1$, otherwise the statement is trivially satisfied. 
This implies none of the states with greater or equal mass decrease to 0 (\cref{thm:2}). Let the smallest non-zero mass state have index $i$. 
From \cref{thm:5}, the decrease in its probability is given by:

\begin{align}
   & \spa \frac{\eta}{c} \spa \bigg (  \bigg (  
   (\sum_j z^t_j) \frac{1}{(\sum_l z^t_l)^2} \sum_k  z^t_k \log \frac{z^t_k}{z^t_i}  \bigg ) - \bigg (  
   z^t_i \sum_j \frac{1}{(\sum_l z^t_l)^2} \sum_k z^t_k \log \frac{z^t_k}{z^t_j}
    \bigg ) \bigg ) \\
    & > \frac{\eta}{c} \spa \bigg (  \bigg (
   (\sum_j z^t_j) \frac{1}{(\sum_l z^t_l)^2} \sum_k  z^t_k \log \frac{z^t_k}{z^t_i} \bigg )  -
   \bigg ( z^t_i \sum_j \frac{1}{(\sum_l z^t_l)^2} \sum_k z^t_k \log \overbrace{\frac{z^t_k}{z^t_i}}^{\text{replaced $z^t_j$}}
    \bigg ) \bigg ) \\
    & = 
   \frac{\eta}{c} \spa \bigg ( (\sum_j z^t_j)  -
   s \spa z^t_i \bigg )   \bigg ( \frac{1}{(\sum_l z^t_l)^2} \sum_k z^t_k \log \frac{z^t_k}{z^t_i}
    \bigg ) \\
    & = \frac{\eta}{c} \spa \bigg ( 1 - s \spa \prob(X = i | z^t) \bigg )  \bigg ( \sum_k  \frac{z^t_k}{\sum_l z^t_l} \log \frac{z^t_k}{z^t_i} \bigg ) \label{eq:lower_bound} \\
    & > 0,
\end{align}
using that $\frac{\eta}{c} > 0$, gradient of smallest non-zero mass is positive (\cref{thm:0b}), and $\prob(X = i | z^t) < \frac{1}{s}$, since $i$ is the smallest element and we assumed the masses are non-uniform (which is reasonable; \cref{thm:3}).
Therefore the decrease in the probability of the smallest element is lower bounded by a positive value. Now we show that this lower bound on $i$'s decrease increases with time (that is, if one assumes the worst case of minimal decrease, it speeds up).
Let $z^t_m$ be the second value in the ascending-sorted vector of non-zero masses in $z^t$.
From the lower bound in \cref{eq:lower_bound}:

\begin{align}
    & \spa \frac{\eta}{c} \bigg ( 1 - s \spa \prob(X = i | z^t) \bigg )  \bigg ( \sum_k  \frac{z^t_k}{\sum_l z^t_l} \log \frac{z^t_k}{z^t_i} \bigg ) \\
    & = \frac{\eta}{c} \bigg ( 1 - s \spa \prob(X = i | z^t) \bigg )  \bigg ( \sum_{k \neq i}  \frac{z^t_k}{\sum_l z^t_l} \log \frac{z^t_k}{z^t_i} \bigg ) \\
    & \geq \frac{\eta}{c} \bigg ( 1 - s \spa \prob(X = i | z^t) \bigg ) \bigg ( \sum_{k \neq i}  \frac{z^t_k}{\sum_l z^t_l} \log \overbrace{\frac{z^t_m}{z^t_i}}^{\text{replaced $z^t_k$}} \bigg ) \\
    & = \frac{\eta}{c} ( 1 - s \spa \prob(X = i | z^t) )  (1 - \prob(X = i | z^t))  \log \frac{z^t_m}{z^t_i}.  \label{eq:four_terms}
\end{align}
Each of the 4 multiplicative factors in \cref{eq:four_terms} is positive and increases with time in any window of timesteps where no mass drops to 0; otherwise, we are done. The first 3 trivially so: $\eta$ is constant, $c$ is the product of old and new total masses and total masses decrease with time (\cref{thm:8}), $s$ is constant, $\prob(X = i | z^t)$ or the probability of the state with smallest mass decreases with time until it becomes 0 (\cref{thm:7,thm:6}). Now we show that the last term, the ratio between the second and first values in the sorted vector of non-zero masses, $z^t_m$ and $z^t_i$, increases with time.

\begin{align}
    \frac{z^t_m - \eta \frac{\df \ent(z^t)}{\df z^t_m}}{z^t_i - \eta \frac{\df \ent(z^t)}{\df z^t_i}} &> \frac{z^t_m}{z^t_i} \\
    \longleftrightarrow - z^t_i \spa \eta \frac{\df \ent(z^t)}{\df z^t_m} &> - z^t_m \spa \eta \frac{\df \ent(z^t)}{\df z^t_i} \\
    \longleftrightarrow z^t_i \spa  \frac{\df \ent(z^t)}{\df z^t_m} &< z^t_m \spa \frac{\df \ent(z^t)}{\df z^t_i} \\
     \longleftrightarrow z^t_i \spa \sum_{j} z^t_j \log \frac{z^t_j}{z^t_m} &< z^t_m \spa \sum_{j} z^t_j \log \frac{z^t_j}{z^t_i} \\
     \longleftrightarrow z^t_i \spa \bigg ((\sum_{j} z^t_j \log z^t_j) - (\sum_{j} z^t_j) \log z^t_m \bigg ) &< z^t_m \spa \bigg ((\sum_{j} z^t_j \log z^t_j) - (\sum_{j} z^t_j) \log z^t_i \bigg ) \\
     \longleftrightarrow (\sum_{j} z^t_j) (z^t_m \log z^t_i - z^t_i \log z^t_m) &< (z^t_m - z^t_i)  (\sum_{j} z^t_j \log z^t_j)\\
     \longleftrightarrow  \frac{z^t_m \log z^t_i - z^t_i \log z^t_m}{z^t_m - z^t_i} &<   \frac{\sum_{j} z^t_j \log z^t_j}{\sum_{j} z^t_j} \\
     \longleftarrow \frac{z^t_m \log z^t_i - z^t_i \log \overbrace{z^t_i}^{\text{replaced } z^t_m}}{z^t_m - z^t_i} &<  \frac{\sum_{j} z^t_j \log z^t_j}{\sum_{j} z^t_j} \\
     \longleftrightarrow  \frac{(z^t_m - z^t_i) \log z^t_i}{z^t_m - z^t_i} &<  \frac{\sum_{j} z^t_j \log z^t_j}{\sum_{j} z^t_j} \\
     \longleftrightarrow & \top.
\end{align}

\hfill$\square$.

\subsection{Expected cardinality of sampled sets}

\begin{figure}[h]
\centering
\includegraphics[width=0.5\textwidth]{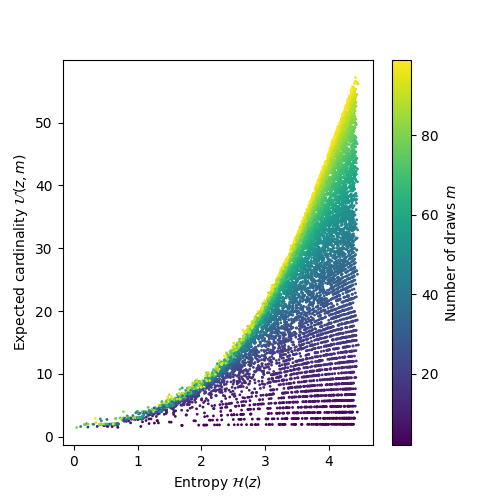}
    \caption{Entropy plotted against expected number of unique states drawn, for 10K samples of masses $z$ with random length $s = |z| \in [2, 100]$ and draws $m \in [2, 100]$.
    %Correlation is stronger for large $m$.
    The upper bound on expected number of unique draws goes up with $m$ because it is not possible to draw more than $m$ unique states if $m$ is the number of draws. 
    }
    \label{fig:full}
\end{figure}  

Now let us consider a different quantity, the expected cardinality or number of unique states obtained with $m$ draws of $X$, $\uniq(z, m)$. 
Let $I_i$ be 1 if we draw at least 1 sample from the $i^{th}$ state of $X$, and 0 if we draw none. Then: 
\begin{align}
    \uniq(z, m) &\triangleq  \ex\big[\sum_i I_i\big] = \sum_i \ex\big[I_i\big] \label{eq:lin} \qquad \text{(linearity of expectation)} \\
    &= \sum_i 0 P(I_i=0 | z) + 1 P(I_i=1 | z) \\
    &= \sum_i 1 - \bigg(1 - \frac{z_i}{\sum_{j} z_j}\bigg)^m,
\end{align}
which is strictly less than the standard cardinality of the set of possible states, $\sum_{i=0}^{s-1} \mathbbm{1}_{z_i > 0}$. The gradient with respect to each mass is:
\begin{align}
  &\frac{d \uniq(z, m)}{d z_k} = \frac{d}{d z_k} \bigg ( 1 - \bigg(1 - \frac{z_k}{\sum_{j} z_j}\bigg)^m \bigg ) + \sum_{i \neq k} \frac{d}{d z_k} \bigg (  1 - \bigg(1 - \frac{z_i}{\sum_{j} z_j}\bigg)^m \bigg )\\
  &= -m  \bigg(1 - \frac{z_k}{\sum_{j} z_j}\bigg)^{m-1} \frac{d}{d z_k} \bigg ( -\frac{z_k}{\sum_{j} z_j} \bigg ) + \sum_{i \neq k} -m  \bigg(1 - \frac{z_i}{\sum_{j} z_j}\bigg)^{m-1}  \frac{d}{d z_k} \bigg ( - \frac{z_i}{\sum_{j} z_j} \bigg ) \\
    &= - m  \bigg(1 - \frac{z_k}{\sum_{j} z_j}\bigg)^{m-1} \frac{z_k - \sum_{i} z_i}{\big ( \sum_{j} z_j \big )^2}
    + \sum_{i \neq k} -m  \bigg(1 - \frac{z_i}{\sum_{j} z_j}\bigg)^{m-1} \frac{z_i}{\big ( \sum_{j} z_j \big )^2}
    \\
    &= \frac{m}{\big ( \sum_{j} z_j \big )^2}  
    \sum_i z_i \bigg ( \bigg(1 - \frac{z_k}{\sum_{j} z_j}  \bigg)^{m-1}  -  \bigg(1 - \frac{z_i}{\sum_{j} z_j}  \bigg)^{m-1} \bigg ). \label{eq:u_grad}
\end{align}

\begin{figure}[h]
    \centering
    \includegraphics[width=0.3\textwidth]{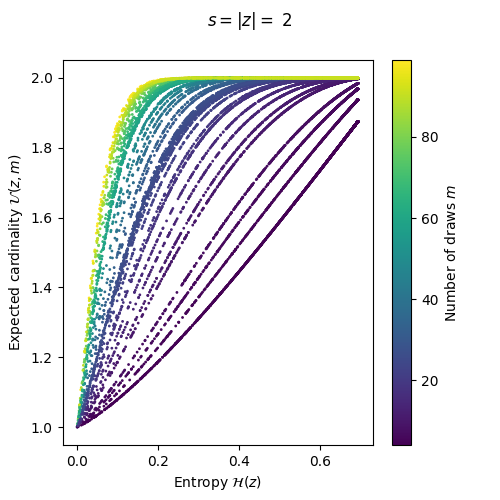}%
    \includegraphics[width=0.3\textwidth]{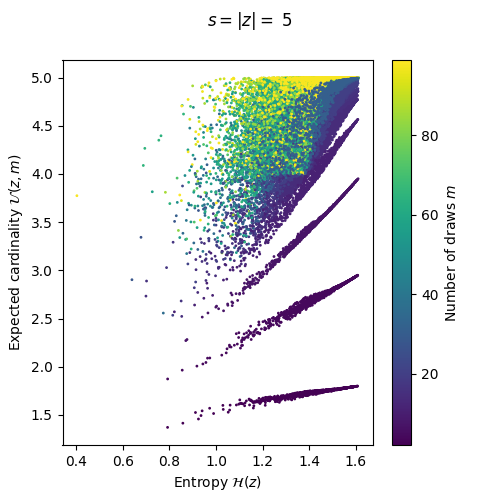}%
    \includegraphics[width=0.3\textwidth]{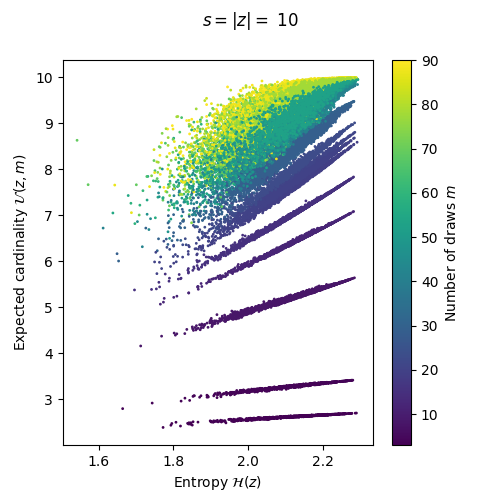}
    
    \includegraphics[width=0.3\textwidth]{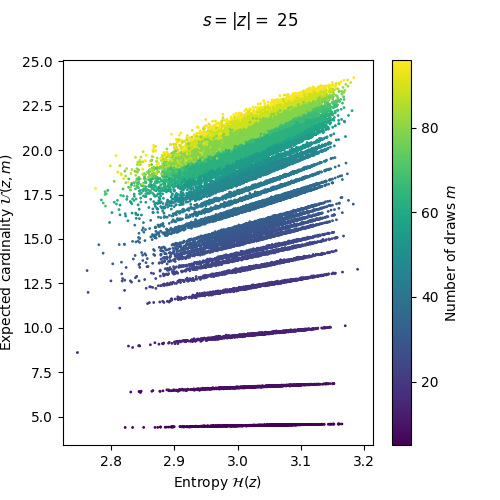}%
    \includegraphics[width=0.3\textwidth]{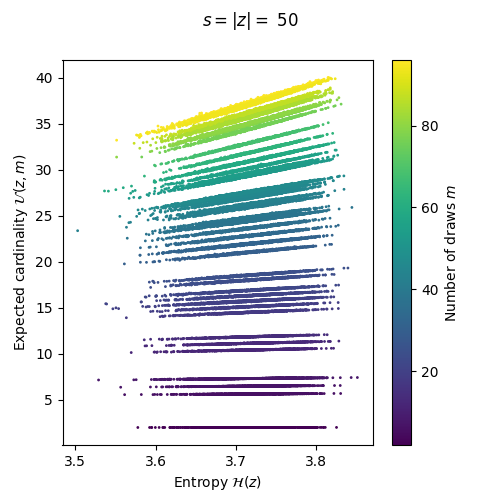}%
    \includegraphics[width=0.3\textwidth]{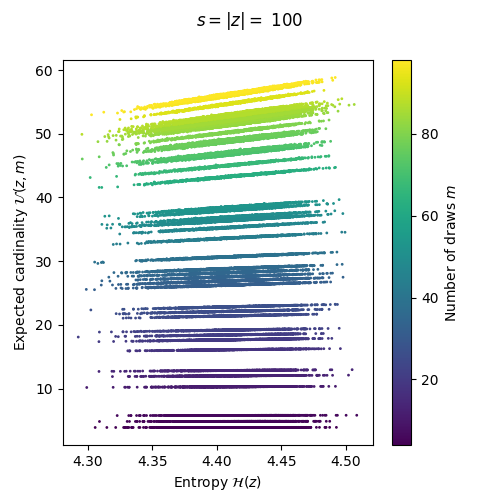}
    \label{fig:breakdown}
    \caption{Partial breakdown of \cref{fig:full} grouped by $s = |z|$, the length of the mass vector or maximum number of possible states. Reducing entropy is not a sufficient condition for reducing expected cardinality in general, given fixed $s$ and $m$, but we show that local optimization via gradient descent on entropy with respect to probability masses is sufficient.}\label{f:breakdown}
\end{figure}

\noindent Entropy $\ent$ and expected number of unique state draws $\uniq$ are related. This is suggested by the similar forms of their gradients (\cref{eq:h_grad2,eq:u_grad}), and also empirically (\cref{fig:full}). A breakdown of \cref{fig:full} into different fixed lengths of $z$ is in \cref{f:breakdown}.
To understand the relationship between the two quantities, note that one intuitively expects higher entropy distributions to have higher expected number of unique draws, given some fixed number of draws. For example, for probability distributions [0.5, 0.5] and [0.9, 0.1], observing both states is more likely sooner with the former.
Reducing entropy makes the distribution more skewed (\cref{s:properties}), therefore sampling more unfair, which reduces the expected number of seen states for any number of draws.
%Now we formalize this intuition, proving that local minimization of entropy also reduces the expected number of unique states sampled in any finite number of draws.

\begin{theorem}
\textbf{Gradient descent on entropy $\ent(z)$ decreases the expected number of unique states in $m$ draws, $\uniq(z, m)$, for any fixed $m > 1$.}
\end{theorem}
%
%Let $S$ denote the indices of zero values in $z$, $i \in S \leftrightarrow z_i = 0$. These indices are also 0 in $\tilde{z}$ (\cref{thm:4}).
%Remove these indices from both $z$ and $\tilde{z}$, then sort $z$ and $\tilde{z}$ in ascending order.
%These operations on $z, \tilde{z}$ do not change the values of $\ent(z), \ent(\tilde{z})$, $\uniq(z, m), \uniq(\tilde{z}, m)$, or $\frac{d\ent(z)}{dz_j}$ for $z_j \neq 0$.
%Let $p_i \triangleq \frac{z_i}{\sum_j z_j}$, $q_i \triangleq 1 - p_i$ and similarly let $\tilde{p}_i \triangleq \frac{\tilde{z}_i}{\sum_j \tilde{z}_j}$ and $\tilde{q} \triangleq 1 - \tilde{p}_i$.

\noindent Without loss of generality since $\ent(z)$ and $\uniq(z, m)$ are invariant to order of $z$, assume $z$ is sorted in ascending order, and therefore also $\tilde{z}$ (\cref{thm:2}).
Define $p_i \triangleq \frac{z_i}{\sum_j z_j}$, $q_i \triangleq 1 - p_i$ and similarly $\tilde{p}_i \triangleq \frac{\tilde{z}_i}{\sum_j \tilde{z}_j}$ and $\tilde{q} \triangleq 1 - \tilde{p}_i$.

When $m=1$: $\uniq(z, 1) - \uniq(\tilde{z}, 1) = 0 \longleftrightarrow \sum_i \tilde{q}_i - q_i = \sum_i (1 - \tilde{p}_i) - (1 - p_i) = 0 \leftrightarrow \top$.
When $m > 1$, $\uniq(z, m)$ decreases during gradient descent on $\ent(z)$ iff:
\begin{align}
    \uniq(z, m) - \uniq(\tilde{z}, m) &> 0 \label{eq:1} \\
    \longleftrightarrow \sum_i 1 - q_i^m - (1 - \tilde{q}_i^m) &> 0 \\
    \longleftrightarrow \sum_i \tilde{q}_i^m &> \sum_{i} q_i^m  \label{eq:2}  \\
    \longleftarrow \sum_{i} \tilde{q}_i^m  > \sum_{i} q_i \spa \tilde{q}_i^{m-1} &\land \sum_{i} q_i \spa \tilde{q}_i^{m-1} > \sum_i q_i^m. \label{e:two_terms}
\end{align}
Now we prove the truth of both conditions in \cref{e:two_terms}. For the first condition in \cref{e:two_terms}:
\begin{align}
    \sum_{i} \tilde{q}_i^m & > \sum_{i} q_i \spa \tilde{q}_i^{m-1} \\
    \longleftrightarrow
    \sum_{i} \tilde{q}_i \spa \tilde{q}_i^{m-1} & > \sum_{i} q_i \spa \tilde{q}_i^{m-1} \\
    \longleftrightarrow
    \sum_{i} \tilde{q}_i^{m-1} \spa (\tilde{q}_i  - q_i) &> 0. \label{e:part1}
\end{align}
Since masses $z$ and $\tilde{z}$ are sorted in ascending order, so are probabilities $p$ and $\tilde{p}$. So $q$ and $\tilde{q}$ are both sorted in decreasing order. The vector of changes in state probabilities $\tilde{p} - p$
%can be split into two contiguous parts $\triangle_p[0..j)$ and $\triangle_p[j, |z|)$ for some $j \in [0, |z|)$ such that $j$ is the index of the first non-zero element in $z$. Then $\sum_{i < j} \tilde{q}_i^{m-1} \spa (\tilde{q}_i  - q_i) = 0$ because masses at indexes smaller than $j$ remain 0 (\cref{thm:4}). Now we show $\sum_{i \geq j} \tilde{q}_i^{m-1} \spa (\tilde{q}_i  - q_i) > 0$.
consists of a sequence of non-positive values followed by a sequence of positive values, and sums to 0. Then $\tilde{q} - q = (1 - \tilde{p}) - (1 - p) = - (\tilde{p} - p)$ 
%$\triangle_q[j, |z|)= - \triangle_p[j, |z|)$ so 
consists of a sequence of non-negative values followed by a sequence of negative values, and also sums to 0.
Since $\tilde{q}$ is soted in decreasing order, so is $\tilde{q}^{m-1}$. Therefore by \cref{thm:gen_sum}, \cref{e:part1} is true. 

For the second condition in \cref{e:two_terms}. Let $\mu \triangleq \frac{1}{s} \sum_i \tilde{q}_i^{m-1} - q_i^{m-1}$:
\begin{align}
    \sum_{i} q_i \spa \tilde{q}_i^{m-1} & > \sum_{i} q_i^m \\
    \longleftrightarrow 
    \sum_{i} q_i \spa \tilde{q}_i^{m-1} & > \sum_{i} q_i \spa q_i^{m-1} \\
    \longleftrightarrow 
    \sum_{i} q_i ( \tilde{q}_i^{m-1} - q_i^{m-1}) & > 0 \\
    \longleftrightarrow 
    \bigg ( \sum_{i} q_i ( \tilde{q}_i^{m-1} - q_i^{m-1} - \mu)  \bigg) +  \mu\sum_{i} q_i & > 0. \label{eq:entropy_uniq_last}
\end{align}
The first term in \cref{eq:entropy_uniq_last} is positive by \cref{thm:gen_sum}. Second term is $\geq 0$ because $\sum_{i} q_i$ is positive and $\mu$ is positive or zero (proof by induction, i.e. assuming  \cref{eq:1} and therefore \cref{eq:2} for $m-1$).
\hfill$\square$.

\section{Continuous variable case}

In continuous spaces, a standard distribution-invariant measure for size is covering number, which is the cardinality of the smallest set of partitions that cover the space, for a given distance metric $\rho$ and upper bound on partition width $\epsilon$.

\begin{definition}[\citet{van1996weak}]For a metric space $S, \rho$ and
$T\subset S$, $\hat{T}\subset S$ is an {\em
$\epsilon$-cover} of $T$, if $\forall t\in T$, $\exists \hat{t}\in
\hat{T}$ such that $\rho(t, \hat{t})\leq \epsilon$. The {\em
$\epsilon$-covering number} of $T$ is
\[\mathcal{N}(\epsilon, T, \rho)=\min\{|\hat{T}|\,: \hat{T} \mbox{ is an }\epsilon-\mbox{cover of }T\}.\]
\end{definition}

We provide two arguments for the link between minimizing entropy and covering number for probability distributions defined on continuous variables.
First, differential entropy is the limiting case of discrete entropy: $\int_x \prob(x) \log \prob(x) \spa dx = \lim_{s \rightarrow \infty} \sum_{i=0}^{s-1} ( \prob(x_i) \log \prob(x_i)) \triangle x $, and the arguments in \cref{s:properties} hold given any discretization of the continuous support of $\mathcal{X}$ into finite partitions. 

Secondly, the relationship is clear to see for specific cases of continuous probability distributions where entropy can be analytically computed. In the case of the uniform distribution defined on the continuous range of scalars $[a, b]$, entropy is $\ln (b - a)$, and since $\ln$ is monotonically increasing, decreasing entropy corresponds to monotonically decreasing the range, which reduces the $\epsilon$-covering number in a finite number of timesteps assuming a positive lower bound for the reduction in range in each optimization timestep.
In the case of the independent multivariate Gaussian distribution on variable $X \in \reals^k$ (i.e. the covariance matrix is diagonal and components are independent), entropy is given by $\frac{1}{2} \sum_{i=0}^{k-1} \ln (2 \pi e \operatorname{Var}[X_i])$ \citep{norwich1993entropy}, so gradient descent on entropy with respect to the variance of each component implies a monotonic decrease in the latter. The smaller the variance of a Gaussian, the sharper it becomes, and the fewer the number of covering partitions with non-negligible probability mass, given any partitioning of the unbounded support of the distribution.

In practice, minimizing entropy of a data representation $X \in \reals^k$ (which is suggested by information bottleneck, \citet{ahuja2021invariance,tishby2000information}) can be implemented as minimizing the empirical variance of each of the $k$ components in the representation vector. The rationale is that the multivariate normal distribution is the maximum entropy distribution for a given covariance, the entropy of the multivariate normal distribution is upper bounded by entropy of the independent variable case of the multivariate normal distribution \citep{kirsch2020unpacking}, and the entropy of the latter is a sum of logs of the variance of each component variable, as we have seen. Gradient descent on the empirical variance with respect to the representation of each sample corresponds to moving the representations closer to the mean, as $\frac{d \sigma^2}{d x_k} = \frac{d \frac{1}{n} \sum_{i=0}^{n-1} (x_i - \bar{x})^2}{d x_k} = \frac{2}{n} (x_k - \bar{x})$, which contracts the space. 

\paragraph{Acknowledgements.} We are grateful for discussions with Razvan Pascanu, Aristide Baratin, and Yoshua Bengio. The first author is supported by funding from IVADO. 

\bibliography{main}
\bibliographystyle{plainnat}

\newpage
\appendix

\section{Lemmas}
\begin{lemma}\label{app:1}
\textbf{For any $a \in \reals_{\geq 0}^s$ of non-identical values, $\sum_j \sum_i a_i \log \frac{a_i}{a_j} > 0$.}
\end{lemma}

We prove this by induction. Without loss of generality since order is immaterial for the sums in \cref{app:1}, assume $a$ is sorted in ascending order. The base case, considering only the first 2 elements of $a$, $a_1 \geq a_0$, is given by:

\begin{align}
    a_0 \log \frac{a_0}{a_1} + a_1 \log \frac{a_1}{a_0} + a_0 \log \frac{a_0}{a_0} + a_1 \log \frac{a_1}{a_1} &\geq 0 \\
    \longleftrightarrow  a_0 \log \frac{a_0}{a_1} + a_1 \log \frac{a_1}{a_0}  &\geq 0 \\
    \longleftrightarrow  a_1 \log \frac{a_1}{a_0}  &\geq - a_0 \log \frac{a_0}{a_1}  \\
    \longleftrightarrow  \frac{- a_1}{a_0}  &\leq   \frac{\log \frac{a_0}{a_1} }{\log \frac{a_1}{a_0} }\\
    \longleftrightarrow  \frac{a_1}{a_0}  &\geq  \frac{\log \frac{a_1}{a_0} }{\log \frac{a_1}{a_0} }\\
    \longleftrightarrow &\top,
\end{align}
where the inequalities are strict iff $a_1 > a_0$.
The inductive case, considering the first $k$ elements and assuming $\sum_{j=0}^{k-1} \sum_{i=0}^{k-1} a_i \log \frac{a_i}{a_j} \geq 0$ holds for the first $k-1$ elements:
\begin{align}
    \sum_{j=0}^k \sum_{i=0}^k a_i \log \frac{a_i}{a_j} &\geq 0 \label{app:lemma1_0} \\
    \longleftrightarrow a_k \log \frac{a_k}{a_k} +  \bigg ( \sum_{i=0}^{k-1} a_i \log \frac{a_i}{a_k} + a_k \log \frac{a_k}{a_i} \bigg ) + \bigg ( \sum_{j=0}^{k-1} \sum_{i=0}^{k-1} a_i \log \frac{a_i}{a_j} \bigg ) & \geq 0 \label{app:lemma1_1} \\
    \longleftarrow \sum_{i=0}^{k-1} a_i \log \frac{a_i}{a_k} + a_k \log \frac{a_k}{a_i} &\geq 0 \\
    \longleftrightarrow \sum_{i=0}^{k-1}  a_i \log a_i + a_k \log a_k  - a_i \log a_k - a_k \log a_i &\geq 0 \\
    \longleftrightarrow \sum_{i=0}^{k-1} (a_k - a_i) \log a_k - (a_k - a_i)\log a_i &\geq 0 \\
    \longleftrightarrow \sum_{i=0}^{k-1} (a_k - a_i) (\log a_k - \log a_i) &\geq 0 \\
    \longleftrightarrow &\top,
\end{align}
%\cref{app:lemma1_0,app:lemma1_1}
where the inequalities are strict if $a_{0\dots k}$ is not a vector of identical values.
\hfill$\square$.
\newpage

\begin{lemma} \label{thm:gen_sum}
\textbf{Let $w \in \mathcal{R}_{\ge 0}^s$ be any non-uniform vector of non-negative reals sorted in descending order, so $\forall i: w_{i} \geq w_{i+1}$. Let $x \in \mathcal{R}^s$ be any non-uniform vector of reals that sums to 0 and consists of a contiguous sequence of non-negative values followed by a contiguous sequence of negative values, so $\sum_i x_i = 0$ and $\exists l \in [1, |x|-1]$ such that $ \forall i \in [0, l): x_i \geq 0$, $\forall i \in [l, |x|): x_i < 0$. Then dot product $w^\intercal x > 0$ is positive.}
\end{lemma}

Let $A = [0, l)$ be the indices of all non-negative elements in $x$ and $B = [l, |x|)$ be the indices of all negative elements in $x$.
\begin{align}
    \sum_i w_i \spa x_i &>0 \\
    \longleftrightarrow \bigg ( \sum_{a \in A} w_a \spa x_a \bigg ) + \bigg ( \sum_{b \in B} w_b \spa x_b \bigg ) &> 0 \\
    \longleftrightarrow 
     \overbrace{\bigg ( \sum_{a \in A} x_a\bigg ) }^{g_A > 0}  \overbrace{\bigg (  \frac{\sum_{a \in A} w_a \spa x_a}{\sum_{a \in A} x_a}\bigg ) }^{f_A > 0}  + 
    \overbrace{\bigg ( \sum_{b \in B} x_b \bigg )}^{g_B < 0}   \overbrace{ \bigg ( \frac{\sum_{b \in B} w_b \spa x_B}{\sum_{b \in B} x_b} \bigg ) }^{f_B > 0}  &> 0 \\
    \longleftrightarrow 
    f_A \spa g_A &> - f_B \spa g_B \\
    \longleftrightarrow 
    \frac{f_A}{f_B} &> - \frac{g_B}{g_A} \\
    \longleftrightarrow 
    \frac{f_A}{f_B} &> \frac{g_A}{g_A} = 1 \\
    \longleftrightarrow  \top,
\end{align}
because $g_A + g_B = 0$ and $f_A > f_B$.
\hfill$\square$.

\end{document}